\documentclass[11pt]{article}

\usepackage[final]{acl}
\usepackage{times}
\usepackage{latexsym}

\usepackage[T1]{fontenc}

\usepackage[utf8]{inputenc}

\usepackage{microtype}

\usepackage{inconsolata}

\usepackage{graphicx}

\usepackage{algorithm}
\usepackage{url}
\usepackage{booktabs}
\usepackage{multirow}
\usepackage{array}
\usepackage{subcaption}
\usepackage{longtable}
\usepackage{marvosym}
\usepackage{amsfonts,bm}
  \usepackage{amsmath,amssymb}   
  \usepackage{algpseudocode}     
\usepackage{stfloats}
\usepackage{tabularx}
\usepackage{caption}
\usepackage{wasysym}
\usepackage{pifont}
 \usepackage{fvextra}
  \DefineVerbatimEnvironment{PromptBlock}{Verbatim}{
    breaklines=true,
    breakanywhere=true,
    fontsize=\footnotesize
  }

\usepackage{xcolor}
  \usepackage{fvextra}
  \DefineVerbatimEnvironment{BlueBlock}{Verbatim}{
    breaklines=true,
    breakanywhere=true,
    fontsize=\scriptsize,
    formatcom=\color{blue}
  }

\title{Pareto-Guided Teacher Alignment for Fair Personalized Text Generation}

\author{
  Tunazzina Islam \\
  Department of Computer Science \\
  Purdue University, West Lafayette \\
  Indiana 47907, USA \\
  \texttt{islam32@purdue.edu} \\}
  
\begin{document}
\maketitle
\begin{abstract}

Personalized persuasive text generation can improve relevance and engagement, but demographic conditioning may also introduce unequal framing across groups. We study fairness mitigation in personalized generation as a constrained multi-objective alignment problem: reduce demographic disparities while preserving personalization fidelity. We propose a \textbf{Pareto-guided teacher alignment} framework that combines revision-based candidate generation, pair-aware feasibility gating, Pareto-style candidate selection, and optional preference optimization through supervised fine-tuning and direct preference optimization. We evaluate the framework on climate change and vaccination persuasion tasks using a controlled context-rich demographic grid with matched gender and age pairs and a unified \textit{five-audit} evaluation suite spanning persuasion bias, formality disparity, emotional framing disparity, lexical association disparity, and personalization fidelity. Across both domains and cross-family transfer settings, no single alignment strategy dominates all objectives simultaneously. Instead, methods occupy different regions of a fairness--personalization Pareto frontier: some achieve stronger disparity reductions, while others better preserve personalization or demographic stability. Our results show that fairness mitigation effects are objective-dependent and transfer inconsistently across domains and model families, motivating bounded-regression, multi-audit model selection over single-metric optimization for fairness-sensitive personalized generation.

  
\end{abstract}

\section{Introduction}

Large language models (LLMs) \cite{brown2020language} are increasingly used to generate personalized persuasive text by conditioning on demographic and contextual attributes such as age group, gender, region, and topic preferences \citep{breum2024persuasive,wang2019persuasion}. In applications such as climate communication, public-health messaging, and civic engagement, personalization can improve relevance and persuasive effectiveness by adapting language to audience-specific contexts \cite{islam2025post,bai2025llm,karinshak2023working}. However, demographic conditioning also introduces a fairness risk: models may vary not only topical emphasis, but also tone, emotional framing, certainty, formality, and persuasive pressure across demographic groups \citep{islam2026gets,borah-etal-2026-persuasion,pauli2026analysing,liu2025llm,tan2025unmasking,wan2023kelly}. Such disparities are often subtle and difficult to detect through aggregate quality metrics alone, particularly in personalized generation settings where stylistic variation is expected.

A central challenge is that fairness mitigation in generation systems is frequently evaluated using a single target metric or isolated bias criterion \citep{wan2025white}. In practice, this can produce misleading conclusions. Reducing one form of demographic disparity does not necessarily improve others, and aggressive mitigation may weaken personalization fidelity or persuasive relevance \citep{kleinberg2016inherent}. Existing mitigation approaches span data augmentation \citep{dinan2020queens}, inference-time control \citep{krause2021gedi,schick2021self}, post-hoc rewriting \citep{wan2025white,stahl2022prefer}, and aggregation-based methods \citep{abels2025wisdom}. While these methods can improve selected metrics, prior work has shown that mitigation effects often transfer inconsistently across tasks, models, and fairness criteria \citep{jin2021transferability}. This motivates a broader perspective in which fairness is treated as a constrained multi-objective alignment problem rather than a single-score optimization problem.
\begin{figure*}
\includegraphics[width=.95\textwidth]{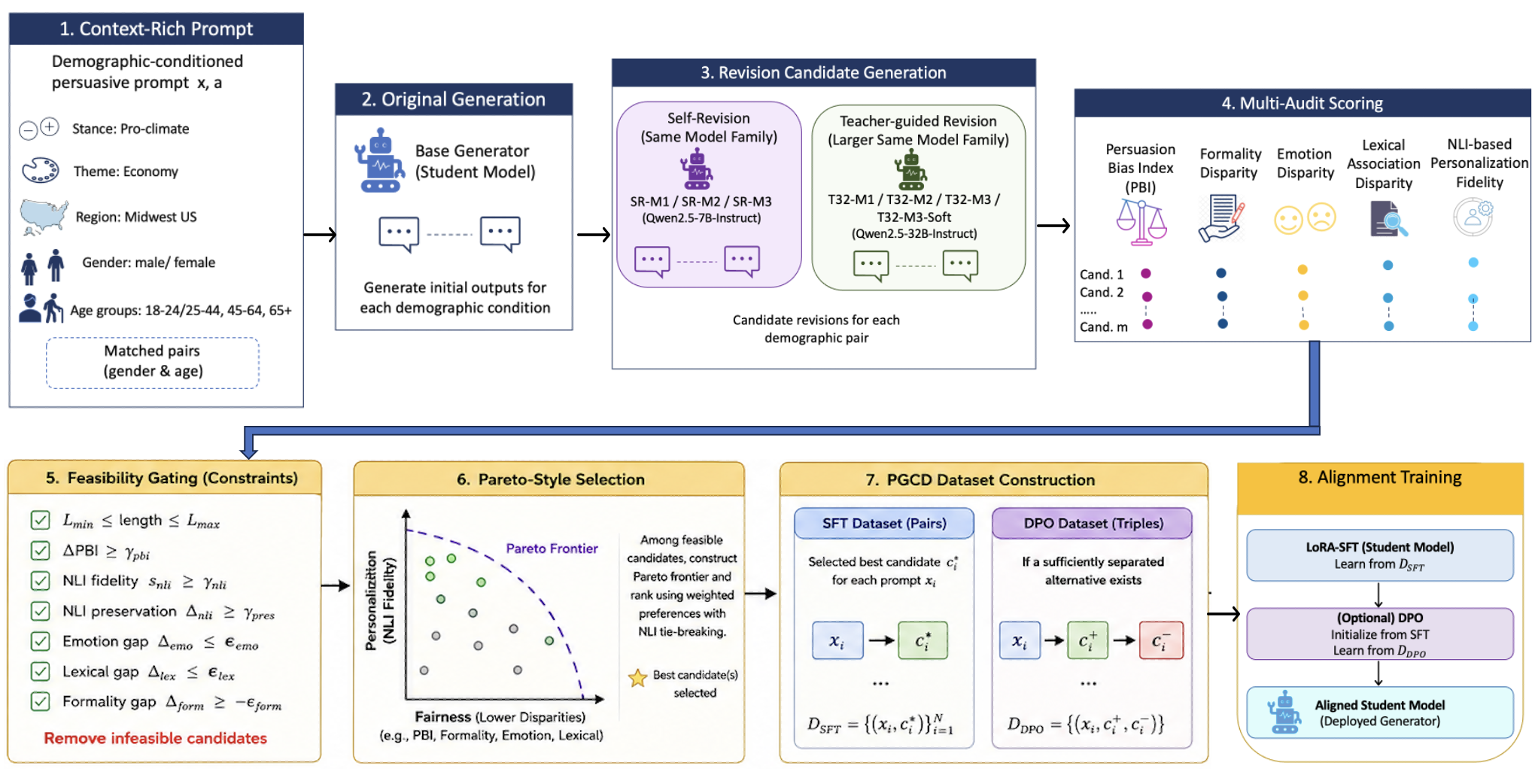}
\caption{{\small 
Overview of the proposed framework.}}
\vspace{-5pt}
\label{fig:frame}
\vspace{-5pt}
\end{figure*}

  
In this work, we study demographic-conditioned personalized text generation under matched-pair evaluation settings, where non-demographic context is fixed and only demographic attributes vary. We formulate alignment as optimizing multiple fairness objectives while preserving personalization fidelity. Concretely, we define a personalization utility objective together with a set of demographic disparity objectives spanning pairwise persuasion bias index (PBI) introduced by \citet{islam2026gets}, formality disparity, emotional framing disparity, lexical association disparity, and personalization fidelity based on natural language inference (NLI) \cite{maccartney2009extended,condoravdi2003entailment,fyodorov2000natural}. Our goal is not to optimize a single fairness score, but to reduce multiple forms of demographic disparity while maintaining high-quality personalized generation.

To address this problem, we propose a \textbf{Pareto-guided teacher alignment} framework for fair personalized generation. The framework combines: (1) revision-based candidate generation using self-revision or larger teacher-guided revision models, (2) pair-aware feasibility gating under explicit fairness and personalization constraints, and (3) Pareto-based candidate selection for constructing aligned supervision and preference data. The resulting candidates are used to train aligned student models through supervised fine-tuning (SFT) and optional direct preference optimization (DPO) \cite{rafailov2023direct}. Unlike prior approaches that optimize narrow message-level objectives, our framework explicitly models fairness--personalization trade-offs during candidate construction and alignment. Fig. \ref{fig:frame} shows the overview of our proposed framework.


 We evaluate the proposed framework on two domains: climate communication and vaccination messaging. Using a controlled demographic grid with matched gender and age pairs, we audit all systems using a unified \textit{five-audit} evaluation suite covering persuasion bias, formality, emotion, lexical associations, and NLI-based personalization fidelity. Across both domains, we find that no single method dominates all objectives simultaneously. Instead, different alignment strategies occupy different regions of a fairness--personalization Pareto \cite{stiglitz1981pareto} frontier: some methods achieve stronger disparity reduction, while others better preserve personalization quality and demographic consistency. These findings suggest that fairness-sensitive generation systems should be selected using bounded-regression, multi-audit criteria rather than single-metric optimization.
 
Our contributions\footnote{Our code and data are publicly available \href{https://github.com/tunazislam/Pareto-Guided-Teacher-Alignment}{github.com/tunazislam/Pareto-Guided-Teacher-Alignment}}
are as follows:\vspace{-0.5em}
\begin{itemize}\setlength{\itemsep}{0pt}\setlength{\parskip}{0pt}
\item We formulate fair personalized persuasive generation as a constrained multi-objective alignment problem balancing demographic fairness and personalization fidelity.

\item We propose a Pareto-guided teacher alignment framework that integrates pair-aware candidate construction, feasibility gating, Pareto filtering, and optional preference optimization through SFT$\rightarrow$DPO training.

\item We introduce a controlled cross-domain evaluation protocol with matched demographic pairing and a unified five-audit fairness suite.

\item We show empirically that fairness mitigation exhibits Pareto-style trade-offs across demographic disparity and personalization objectives, motivating bounded-regression model selection instead of single-metric ranking.

\end{itemize}
  
   

 \section{Related Work}
  \paragraph{Bias and Fairness in LLM generation.}
  A growing literature documents demographic disparities in LLM outputs, including differences in wording, framing, and social tone across groups \cite{wang2026invests,islam2026gets,pauli2026analysing,pauli2025measuring,sethi2025ai,tan2025unmasking,aghaebe2025llms,fang2024bias,dai2024bias,kumar2023language,li2023fairness,li2023survey,sag2023fairness,urman2023silence,kotek2023gender,esiobu2023robbie,wan2023kelly}. Prior
  studies show that these effects appear in both open-ended and task-specific generation, and can be amplified in demographic-conditioned prompting \cite{islam2026gets,wan2023kelly}. Our work
  builds on this line by focusing on persuasive personalization, where fairness and relevance objectives are tightly coupled.
 \paragraph{Persuasive text generation.}
 Persuasive text generation is an active NLP area, with prior work examining how models use argumentation, emotion, and moral framing to influence readers \cite{breum2024persuasive,karinshak2023working,wang2019persuasion}. However, relatively little work studies whether these
  persuasive strategies shift systematically across demographic groups, or how demographic personalization interacts with persuasive framing disparity.  This gap is central to our setting: our task is demographic-conditioned persuasive generation, where relevance and fairness can conflict. Accordingly, we evaluate not only persuasive quality but also demographic disparity across matched pairs, and optimize alignment with
  pair-aware SFT/DPO under multi-audit constraints.

  \paragraph{Bias mitigation across the pipeline.}
  Existing mitigation methods span data-level interventions (e.g., counterfactual augmentation \cite{dinan2020queens}), model-level adaptation (including parameter-efficient fine-
  tuning \cite{masoudian-etal-2024-effective}), and inference-time control (e.g., decoding constraints, filtering, re-ranking \cite{krause2021gedi,schick2021self}), aggregation across multiple outputs \citep{abels2025wisdom}, distribution-level adjustments using counterfactual contexts \cite{banerjee2024all},  post-hoc selective rewriting \cite{wan2025white}. These methods often improve selected metrics, but improvements may not transfer reliably across tasks or fairness criteria \cite{jin2021transferability}. We complement prior work by using a single student family and controlled settings to isolate the effect of alignment strategy and acceptance policy.
 A smaller body of work studies post-hoc rewriting of generated text to reduce bias while preserving meaning, including lexical and instruction-based edits \cite{wan2025white, stahl2022prefer}. A recurring challenge in this setting is \emph{objective mismatch}:
  optimizing a narrow, message-level bias metric does not necessarily improve broader pairwise or distributional fairness criteria. Motivated by this gap, we move from post-hoc black-box mitigation to training-time alignment (SFT and DPO), where pair-aware candidate
  construction and feasibility gating are used to improve fairness--personalization trade-offs under explicit constraints.





  \paragraph{Counterfactual evaluation and fairness trade-offs.}
  Counterfactual fairness perspectives motivate matched-pair evaluation, where non-demographic context is fixed and only demographic attributes vary \cite{zhou2024counterfactual,rosenblatt2023counterfactual,kusner2017counterfactual}. More broadly, fairness theory shows that different fairness notions can be incompatible, so improvements on one objective may worsen another \cite{chouldechova2017fair,kleinberg2016inherent}. Our evaluation
  protocol operationalizes this directly by reporting multiple disparity families together with personalization fidelity, and by interpreting results as
  Pareto-style trade-offs rather than a single best score.

  Relative to prior work, this paper contributes an auditable, cross-domain alignment workflow for personalized persuasion and an empirical analysis of
  fairness--personalization trade-offs under controlled demographic pairing. The goal is not to claim universal fairness, but to provide a reproducible method
  for selecting models under explicit multi-objective constraints.

\section{Problem Formulation}
  Let $x$ denote a non-demographic context (domain, stance, region, theme, and prompt metadata), and let $a$ denote a demographic attribute value.
  A generator $G_{\theta}$ produces a persuasive response \vspace{-0.5em}
  \[
  y = G_{\theta}(x, a).
  \]
  We study two demographic axes, $z \in \{\text{gender}, \text{age}\}$, and construct matched pairs by fixing $x$ and changing only $a$ along one axis.

  Let $\mathcal{D}=\{(x_i,a_i)\}_{i=1}^{N}$ be the context grid, and let $\mathcal{P}_z$ be matched pairs for axis $z$.
  Our goal is to reduce demographic disparity while preserving personalization quality.

  \paragraph{Personalization utility.}
  For each example, we compute instance-level personalization fidelity $u_i(\theta)$ (NLI-based). The mean utility is 
  \vspace{-0.5em}
  \[
  P(\theta)=\frac{1}{N}\sum_{i=1}^{N}u_i(\theta),
  \] \vspace{-0.5em}
  where higher is better.

  \paragraph{Fairness objectives.}
  For each audit family $k$ (PBI, formality, emotion, lexical, and personalization-gap), we compute disparity $d^{(k)}(\theta)$ on matched pairs or group-
  conditioned aggregates, and define \vspace{-0.5em}
  \[
  F_k(\theta)=d^{(k)}(\theta), \qquad k=1,\dots,K,
  \]
  with lower values indicating lower disparity.

  \paragraph{Multi-objective target.}
  We use a constrained objective: \vspace{-0.5em}
  \[
  \min_{\theta}\; \big(F_1(\theta),\dots,F_K(\theta)\big)
  \quad \text{s.t.}\quad P(\theta)\ge \tau .
  \]
Because these objectives are not jointly optimized through a single differentiable loss and no single model is guaranteed to dominate all objectives simultaneously, we operationalize alignment through candidate-level constrained selection and evaluate systems using Pareto-style trade-offs across audits.

\section{Method}

We study fairness mitigation for demographic-conditioned persuasive generation through a pair-aware alignment pipeline. The proposed framework combines revision-based candidate generation, feasibility-constrained filtering, Pareto-style candidate selection, and optional preference optimization. We compare two settings: (i)\textbf{ self-revision baselines} using the same student model family and (ii) \textbf{same-family teacher-guided alignment} using a \textit{larger} teacher model.

\subsection{Overview and Variant Families}  

We use three model families throughout the paper.

\textbf{Original (Orig).}
The unaligned base generator produces persuasive messages directly from demographic-conditioned prompts.

\textbf{Self-revision variants (SR-*).}
Self-revision methods use the same model family for both generation and revision:\vspace{-0.5em}
\begin{itemize}\setlength{\itemsep}{0pt}\setlength{\parskip}{0pt}
\item \textbf{SR-M1}: row-level self-revision SFT,
\item \textbf{SR-M2}: pair-aware PBI-focused SFT,
\item \textbf{SR-M3}: pair-aware multi-objective SFT.
\end{itemize}

\textbf{Teacher-guided variants (T32-*).}
Teacher-guided methods use a larger same-family teacher model to generate revision candidates while keeping the student family fixed:\vspace{-0.5em}
\begin{itemize}\setlength{\itemsep}{0pt}\setlength{\parskip}{0pt}
\item \textbf{T32-M1}, \textbf{T32-M2}, \textbf{T32-M3}: teacher-guided variants of M1/M2/M3,
\item \textbf{T32-M3-Soft}: relaxed multi-objective acceptance,
\item \textbf{T32-Pref-Std}: preference optimization from constrained candidate selection,
\item \textbf{T32-Pref-NLIStable}: preference optimization with NLI-preservation-aware selection and numerically stable DPO training.
\end{itemize}

This design isolates teacher-scale and acceptance-policy effects while keeping the student architecture fixed.

\subsection{Pair-aware Candidate Construction}

For each context \(x_i\), we generate a candidate set 
\vspace{-0.5em}
\[
C_i = \{c_{i1}, \ldots, c_{im}\},
\]
where candidates are obtained through revision prompts applied either independently or jointly across matched demographic pairs.

For preference variants, we build each context-level candidate pool as \vspace{-0.5em}
{\small
  \[
  \mathcal{C}_i=\{y_i^{\mathrm{Orig}},\,y_i^{\mathrm{T32\text{-}M1}},\,y_i^{\mathrm{T32\text{-}M2}},\,y_i^{\mathrm{T32\text{-}
  M3}},\,y_i^{\mathrm{T32\text{-}M3\text{-}Soft}}\}.
  \]
  }
  i.e., the union of outputs from Original and four teacher-guided run families under the same context.
  
Single-message revision methods (M1) revise outputs independently using audit-triggered feedback. Pair-aware methods (M2/M3) jointly revise matched demographic pairs while preserving shared context and reducing demographic framing disparity. Specifically, paired revisions are instructed to maintain comparable tone, certainty, formality, emotional framing, and persuasive pressure across demographic conditions while preserving audience relevance. Each candidate is scored with multiple audit signals:
\vspace{-0.5em}
\begin{itemize}\setlength{\itemsep}{0pt}\setlength{\parskip}{0pt}
    \item \( \Delta_{\text{pbi}} \): pairwise persuasion-bias improvement,
    \item \( \Delta_{\text{form}} \): formality-gap improvement,
    \item \( \Delta_{\text{emo}} \): emotion-gap change,
    \item \( \Delta_{\text{lex}} \): lexical-association change,
    \item \( s_{\text{nli}} \): NLI-based personalization fidelity,
    \item \( \Delta_{\text{nli}} \): NLI preservation relative to the original output.
\end{itemize}

We additionally enforce basic validity constraints including bounded length, non-empty generations, minimum personalization fidelity, and bounded regression thresholds. These candidate-level annotations form the basis for downstream feasibility gating and Pareto-style selection.

\subsection{Feasibility Gating and Pareto-style Selection}
A candidate is considered feasible only if it satisfies all acceptance constraints: 
\vspace{-0.5em}
  \[
  \begin{aligned}
  &L_{\min} \le \text{len}(c_{ij}) \le L_{\max},\  \\
  &\Delta_{\text{pbi}} \ge \gamma_{\text{pbi}}, \quad
  s_{\text{nli}} \ge \gamma_{\text{nli}},\quad
  \Delta_{\text{nli}} \ge \gamma_{\text{pres}},\ \\
  &\Delta_{\text{emo}} \le \epsilon_{\text{emo}},\quad
  \Delta_{\text{lex}} \le \epsilon_{\text{lex}},\quad
  \Delta_{\text{form}} \ge -\epsilon_{\text{form}}.
  \end{aligned}
  \]

These constraints operationalize bounded-regression alignment: candidates must improve targeted fairness objectives without substantially degrading personalization fidelity or introducing regressions on other audit dimensions.

Among feasible candidates, we compute a Pareto-style frontier over the multi-objective audit vectors and select final candidates using weighted ranking with NLI-based tie breaking. Because no single candidate is guaranteed to dominate all objectives simultaneously, selection is treated as a constrained trade-off process rather than a single-score optimization problem.

\subsection{Alignment Dataset Construction}

Selected candidates are converted into alignment datasets for supervised fine-tuning (SFT) and preference optimization.

For supervised alignment, the selected candidate \(c_i^\star\) becomes the target response:
\vspace{-0.5em}
\[
D_{\text{SFT}} = \{(x_i, c_i^\star)\}_{i=1}^{N}.
\]

We train SFT Low-Rank Adaptation (LoRA) \cite{hu2022lora} adapters on \(D_{\text{SFT}}\) to obtain aligned student models.

When sufficiently separated, feasible candidates exist, we additionally construct preference pairs:
\vspace{-0.5em}
\[
D_{\text{DPO}}
=
\{(x_i, c_i^{+}, c_i^{-})\},
\]
where \(c_i^{+}\) is the preferred candidate and \(c_i^{-}\) is a lower-ranked feasible alternative satisfying a minimum margin constraint:
\vspace{-0.5em}
\[
s(c_i^{+}) - s(c_i^{-}) \ge \delta .
\]

DPO preference pairs are sampled only from this fixed multi-run pool after feasibility filtering and margin checks. Preference optimization is initialized from the SFT adapter and trained using conservative DPO settings for numerical stability.  

We refer to the overall alignment pipeline as \textbf{Pair-aware Global-Constrained Distillation (PGCD)}, since aligned supervision and preference datasets are distilled from constrained candidate selection under multiple fairness and personalization objectives. Algorithm~\ref{alg:pgcd} summarizes the complete PGCD pipeline.






\subsection{Inference-time Constraints}
During inference, we enforce bounded rewrite retries and target-length constraints to prevent short-response collapse and degenerate persuasive outputs. This improves audit stability and reduces pathological outputs that can artificially affect fairness metrics.
\begin{algorithm}[t]
\caption{PGCD: Pair-aware Global-Constrained Distillation}
\label{alg:pgcd}
\small
\begin{algorithmic}[1]
\Require Context set \(X=\{x_i\}_{i=1}^{N}\), candidate pool \(C_i\) per context, thresholds \(\Gamma\), weights \(w\)
\Ensure SFT dataset \(D_{\text{SFT}}\), DPO dataset \(D_{\text{DPO}}\)

\State \(D_{\text{SFT}} \gets \emptyset\)
\State \(D_{\text{DPO}} \gets \emptyset\)

\For{each context \(x_i \in X\)}
    \State \(F_i \gets \emptyset\) \Comment{feasible candidates}

    \For{each candidate \(c \in C_i\)}
        \State compute metrics:
        \[
        \Delta_{\text{pbi}},
        \Delta_{\text{form}},
        \Delta_{\text{emo}},
        \Delta_{\text{lex}},
        s_{\text{nli}},
        \Delta_{\text{nli}}
        \]

        \If{\(c\) satisfies all gates in \(\Gamma\)}
            \State add \(c\) to \(F_i\)
        \EndIf
    \EndFor

    \If{\(F_i = \emptyset\)}
        \State continue
    \EndIf

    \State \(P_i \gets \textsc{ParetoFront}(F_i)\)

    \State score each \(c \in P_i\) using weighted objective \(s(c;w)\)

    \State \(c_i^\star \gets \textsc{SelectBestWithNLITiebreak}(P_i)\)

    \State add \((x_i, c_i^\star)\) to \(D_{\text{SFT}}\)

    \State \(c_i^- \gets\) lowest-scoring valid alternative in \(F_i \setminus \{c_i^\star\}\)

    \If{\(c_i^-\) exists and \(s(c_i^\star)-s(c_i^-)\ge\delta\)}
        \State add \((x_i, c_i^\star, c_i^-)\) to \(D_{\text{DPO}}\)
    \EndIf
\EndFor

\State \Return \(D_{\text{SFT}}, D_{\text{DPO}}\)
\end{algorithmic}
\end{algorithm}




\section{Experiments}
  \label{sec:exp}

\subsection{Tasks and Context Grid}
We evaluate two persuasive-generation domains: climate communication and vaccination messaging. Both domains use the same controlled demographic grid under a context-rich prompting setup. Each prompt is defined by \textit{stance} $\times$ \textit{gender} $\times$ \textit{age group} $\times$ \textit{region} $\times$ \textit{theme}, with one sample per cell.
  This yields 240 climate instances ($1 \times 2 \times 4 \times 5 \times 6$) and
  200 vaccination instances ($1 \times 2 \times 4 \times 5 \times 5$), for 440 instances total. In this work, we use only \textbf{pro-climate} and \textbf{pro-vaccination} stances. The pro-climate themes are adopted from \citet{islam2026gets}. We collect pro-vaccination themes from \url{https://www.apha.org/topics-and-issues/vaccines}. We have four age groups adopted from \citet{islam2025post}.
  For fairness auditing, we construct matched demographic pairs within each domain:
  (i) gender pairs (male/female) with all other fields fixed, and
  (ii) age pairs (all age-group combinations) with all other fields fixed.
  This produces 120/100 gender pairs and 360/300 age pairs (climate/vaccination), respectively. Domain-specific themes and shared demographic axes are provided in Table \ref{tab:appendix-domain-axes} in App. \ref{app:axes}.

\subsection{Models}

All aligned systems use the same student family,
\texttt{Qwen/Qwen2.5-7B-Instruct}, trained using LoRA adapters. Teacher-guided variants use
\texttt{Qwen/Qwen2.5-32B-Instruct}
to generate revised candidates through a vLLM OpenAI-compatible API. This setup isolates teacher-scale and alignment-policy effects while keeping the student architecture fixed. Compared systems are given in App. \ref{app:base}.
Prompt templates and examples are provided in App. \ref{app:pt} and \ref{app:pe}. 





\subsection{Training Configuration}

All trainable systems use LoRA adapters on
\texttt{Qwen/Qwen2.5-7B-Instruct}
with rank $r{=}16$, $\alpha{=}32$, dropout $0.05$, target modules \{\texttt{q\_proj}, \texttt{v\_proj}\}. Teacher-guided revision generation uses deterministic decoding:
\texttt{temperature}=0.0 and \texttt{top-p}=1.0. For base generation, we use \texttt{temperature}=0.7 and \texttt{top-p}=0.9. Complete hyperparameters are reported in Tables~\ref{tab:train_hparams} and~\ref{tab:data_hparams} in App. \ref{app:hyper}. We run all experiments on a single-GPU server with the setup provided in Table \ref{tab:compute_env} in App. \ref{app:compute}.

\begin{table*}[t]
  \centering
  \small
  \setlength{\tabcolsep}{2.5pt}
  \resizebox{\textwidth}{!}{
  \begin{tabular}{lrrrrccrrrr}
  \hline
  Run & PBI-G $\downarrow$ & PBI-A $\downarrow$ & Form-G $\downarrow$ & Form-A(F) $\downarrow$ & Emo-G(sig) $\downarrow$ & Emo-A(sig) $\downarrow$ & Lex-G $\downarrow$ & Lex-A $\downarrow$ & NLI-Pers $\uparrow$ & NLI-Pers-Gap $\downarrow$ \\
  \hline
  Orig & 0.8333 & 0.8393 & 0.0713 & 89.26 & 11 & 20 & 0.8998 & 1.1531 & 0.9536 & 0.0091 \\
  SR-M1 & 0.7843 & 0.8480 & 0.0436 & 53.75 & 6 & 23 & 0.4905 & 1.3902 & 0.9508 & 0.0042 \\
  SR-M2 & 0.7914 & 0.7962 & 0.0457 & 44.24 & 2 & \textbf{18} & 0.5936 & 1.0154 & 0.9492 & 0.0064 \\
  SR-M3  & 0.7778 & 0.7815 & 0.0319 & 47.08 & \textbf{1} & 22 & 0.8629 & 1.2953 & 0.9446 & \textbf{0.0005} \\
  T32-M1 & \textbf{0.7472} & 0.7544 & 0.0238 & 46.49 & 2 & 20 & 0.6029 & 0.9462 & 0.9251 & 0.0128 \\
  T32-M2 & 0.8183 & 0.8202 & 0.0147 & 40.72 & \textbf{1} & 21 & 0.5109 & 1.1191 & 0.9293 & 0.0211 \\
  T32-M3 & 0.8769 & 0.8671 & 0.0388 & 45.25 & 3 & 21 & 1.0636 & 1.2394 & \textbf{0.9577} & 0.0060 \\
  T32-M3-Soft & 0.8700 & 0.8200 & 0.0149 & 61.32 & 2 & 22 & 1.0338 & 0.8937 & 0.9454 & 0.0084 \\
  T32-Pref-Std & 0.7594 & \textbf{0.7424} & \textbf{0.0069} & \textbf{31.07} & 4 & 21 & 0.3982 & \textbf{0.5572} & 0.9469 & 0.0178 \\
  T32-Pref-NLIStable & 0.7711 & 0.8081 & 0.0240 & 50.00 & 2 & \textbf{18} & \textbf{0.3616} & 0.6087 & 0.9494 & 0.0110 \\
  \hline
  \end{tabular}
  }
  \caption{{\small \textbf{Climate} domain. Lower is better for fairness/disparity metrics; higher is better for NLI personalization mean. Emo-G(sig)/Emo-A(sig) are counts of significant labels ($p<0.05$). A: Age, G: Gender.}}
  \label{tab:climate_res}
  \end{table*}
\begin{table*}[t]
\centering
\small
\setlength{\tabcolsep}{2.5pt}
\resizebox{\textwidth}{!}{
\begin{tabular}{lrrrrccrrrr}
\hline
Run & PBI-G $\downarrow$ & PBI-A $\downarrow$ & Form-G $\downarrow$ & Form-A(F) $\downarrow$ & Emo-G(sig) $\downarrow$ & Emo-A(sig) $\downarrow$ & Lex-G $\downarrow$ & Lex-A $\downarrow$ & NLI-Pers $\uparrow$ & NLI-Pers-Gap $\downarrow$ \\
\hline
Orig & 0.8437 & 0.9501 & 0.0350 & 65.85 & 1 & 24 & 0.3854 & 0.4965 & 0.9590 & 0.0094 \\
SR-M1  & 0.8173 & 0.9209 & 0.0272 & 48.52 & 1 & 23 & 0.4529 & 0.4064 & 0.9576 & 0.0074 \\
SR-M2 & 0.7640 & 0.9420 & 0.0464 & 35.89 & 1 & 19 & 0.2611 & \textbf{0.3172} & \textbf{0.9632} & 0.0032 \\
SR-M3  & 0.8297 & 0.8986 & 0.0280 & 34.32 & 2 & 24 & 0.7136 & 0.5346 & 0.9565 & 0.0100 \\
T32-M1  & 0.8223 & 0.9257 & 0.0174 & 46.99 & \textbf{0} & \textbf{15} & 0.3294 & 0.4373 & 0.9129 & 0.0271 \\
T32-M2 & 0.8513 & 0.9182 & 0.0567 & 47.76 & \textbf{0} & 24 & \textbf{0.2417} & 0.4484 & 0.9556 & 0.0083 \\
T32-M3  & \textbf{0.7467} & \textbf{0.8642} & \textbf{0.0074} & 46.41 & 2 & 23 & 0.4521 & 0.5772 & 0.9619 & 0.0070 \\
T32-M3-Soft & 0.7527 & 0.8707 & 0.0554 & \textbf{32.04} & 2 & 22 & 0.3447 & 0.3816 & 0.9605 & 0.0095 \\
T32-Pref-Std  & 0.9267 & 0.9558 & 0.0447 & 65.38 & 2 & 24 & 0.4806 & 0.4789 & 0.9503 & 0.0094 \\
T32-Pref-NLIStable  & 0.8720 & 0.9273 & 0.0234 & 54.92 & 1 & 25 & 0.2859 & 0.6084 & 0.9507 & \textbf{0.0021} \\
\hline
\end{tabular}
}
\caption{{\small \textbf{Vaccination} domain. A: Age, G: Gender.}} 
\vspace{-5 pt}
\label{tab:vaccination_res}
\end{table*}  
\subsection{Evaluation Protocol}

All systems are evaluated on the same context grid using a unified \textbf{five-audit} evaluation suite.

\paragraph{PBI Audit.}
We compute pairwise persuasion-bias gaps across matched gender and age pairs following the Persuasion Bias Index (PBI) framework \cite{islam2026gets}. PBI is grounded in established theories of agency and connotation \citep{abele2018agency,sap2017connotation}, we follow the implementation from \citet{islam2026gets} including an agency lexicon, modality and hedging indicators for certainty, dependency-based imperative detection using \texttt{spaCy}.


\paragraph{Formality Audit.}
We evaluate demographic disparities in formality using
\texttt{s-nlp/roberta-base-formality-ranker}  \cite{10.1007/978-3-031-35320-8_4}.
Gender significance uses Welch’s \(t\)-test, while age significance uses one-way ANOVA.

\paragraph{Emotion Audit.}
We analyze emotional framing disparities using
\texttt{SamLowe/roberta-base-go\_emotions}\footnote{\url{https://huggingface.co/SamLowe/roberta-base-go_emotions}}.
We report counts of statistically significant demographic differences across emotion labels.

\paragraph{Lexical Audit.}

We measure lexical disparity with two complementary components. First, we compute salience using odds ratios (OR)
  \cite{szumilas2010explaining} over extracted lexical items (nouns/adjectives), reporting demographic over-/
  under-representation patterns (male vs female; focal age group vs rest). Second, we compute Word Embedding Association Test (WEAT)-style association measures
  \cite{caliskan2017semantics} with predefined stereotype-oriented attribute sets from \citet{islam2026gets,wan2023kelly}. For gender, we use \textit{career} vs.\
  \textit{family} and \textit{power} vs.\ \textit{support}. For age, we use \textit{innovation} vs.\ \textit{tradition} and
  \textit{energy} vs.\ \textit{experience}. We report association effect sizes (and age-group overrepresented-term
  associations) as lexical-bias indicators, where lower cross-group disparity is better.
  
\paragraph{NLI Personalization Audit.}
We evaluate personalization fidelity using
\texttt{MoritzLaurer/deberta-v3-large-zeroshot\\-v2.0} \cite{laurerbuilding2023}.
We report both mean personalization fidelity and demographic consistency gaps.

Lower values indicate lower demographic disparity for fairness metrics, while higher values indicate stronger personalization fidelity. Because objectives may conflict, we interpret results as Pareto-style fairness--personalization trade-offs rather than expecting a single system to dominate all metrics simultaneously.




  
  
  
  
  
  

\section{Results}
\subsection{Overall Trade-offs Across Audits}
Tables~\ref{tab:climate_res} and~\ref{tab:vaccination_res}
show that no single alignment strategy dominates all fairness and personalization objectives simultaneously. Instead, methods occupy different regions of a Pareto-style
fairness--personalization frontier. Teacher-guided variants generally reduce demographic disparities more aggressively, while pair-aware and preference-based variants preserve personalization fidelity and demographic
stability.

\subsubsection{Climate Domain}
In the climate domain (Table~\ref{tab:climate_res}),
disparity reduction is distributed across multiple method families. Among Qwen7B self-revision variants, SR-M3 achieves the strongest pair-aware behavior, including the
best PBI-G (0.7778), PBI-A (0.7815), lowest Form-G (0.0319), and near-zero NLI-Pers-Gap (0.0005). SR-M2 further improves emotional and lexical consistency, achieving the lowest Emo-A(sig) (18) and reduced Lex-A (1.0154).

Teacher-guided alignment further improves several fairness
objectives. T32-Pref-Std achieves the strongest overall
performance on PBI-A (0.7424), Form-G (0.0069), Form-A(F) (31.07), and Lex-A(0.5572), while
T32-M1 achieves the lowest PBI-G (0.7472). These results suggest that teacher-guided candidate construction combined with constrained selection is particularly effective for reducing pairwise persuasion and stylistic disparities.

However, stronger disparity reduction does not uniformly preserve personalization fidelity. T32-M1 substantially
reduces multiple disparity metrics but lowers NLI-Pers to 0.9251, while T32-M3 preserves the highest personalization mean (0.9577) among teacher-guided systems with more
moderate fairness improvements. Preference-based variants improve lexical and formality disparities but may increase
personalization gaps relative to pair-aware SFT-only models.
\vspace{-0.5em}
\subsubsection{Vaccination Domain}
Table~\ref{tab:vaccination_res} shows a similar but domain-dependent trade-off structure for vaccination messaging. Teacher-guided multi-objective alignment is
strongest for core disparity reduction: T32-M3 achieves
the lowest PBI-G (0.7467), PBI-A (0.8642), and Form-G (0.0074), while T32-M3-Soft further improves age-related
formality disparity with the best Form-A(F) (32.04).
T32-M1 achieves the lowest emotional-significance counts
(Emo-G(sig)=0, Emo-A(sig)=15), indicating stronger suppression of demographic emotional framing differences.

At the same time, aggressive mitigation can reduce personalization fidelity and demographic consistency.
T32-M1 lowers NLI-Pers to 0.9129 and increases
NLI-Pers-Gap to 0.0271 despite strong fairness reductions.
In contrast, SR-M2 preserves the highest personalization
mean (0.9632) while also achieving the best Lex-A (0.3172). T32-Pref-NLIStable achieves the lowest NLI-Pers-Gap (0.0021), suggesting that NLI-preservation-aware preference optimization improves demographic consistency
even when core PBI reductions are more moderate.

 \begin{table*}[t]
  \centering
  \small
  \setlength{\tabcolsep}{2.5pt}
  \resizebox{\textwidth}{!}{
  \begin{tabular}{lrrrrccrrrr}
  \hline
  Run & PBI-G $\downarrow$ & PBI-A $\downarrow$ & Form-G $\downarrow$ & Form-A(F) $\downarrow$ & Emo-G(sig) $\downarrow$ & Emo-A(sig) $\downarrow$ & Lex-G $\downarrow$ & Lex-A $\downarrow$ & NLI-Pers $\uparrow$ & NLI-Pers-Gap $\downarrow$ \\
  \hline
  Orig-GM  & 0.9756 & 0.9341 & 0.0511 & 103.85 & 1 & 14 & 0.4911 & 0.5937 & \textbf{0.9852} & 0.0012 \\
  SR-M2-GM  & 0.8257 & 0.8316 & 0.0517 & 24.96 & 1 & 9 & 0.4681 & 0.5340 & 0.9724 & 0.0087 \\
  SR-M3-GM & 0.7940 & 0.8748 & 0.1097 & 34.32 & 1 & 9 & 0.6581 & 0.6384 & 0.9823 & 0.0014 \\
  T32-M2-GM & \textbf{0.7731} & \textbf{0.7695} & 0.0850 & 53.79 & \textbf{0} & 16 & 0.5377 & 0.7418 & 0.9848 & \textbf{0.0008} \\
  T32-M3-GM & 0.8808 & 0.9364 & 0.0989 & 19.45 & 5 & 6 & 0.3601 & 0.4756 & 0.9810 & 0.0067 \\
  T32-M3-Soft-GM  & 0.8756 & 0.8409 & 0.1140 & 25.48 & 6 & \textbf{5} & 0.4460 & \textbf{0.4371} & 0.9815 & 0.0023 \\
  T32-Pref-Std-GM & 0.9407 & 0.9392 & \textbf{0.0479} & \textbf{14.66} & 4 & 9 & \textbf{0.2889} & 0.4916 & 0.9794 & 0.0056 \\
T32-Pref-NLIStable-GM & 0.9006 & 0.8617 & 0.0612 & 17.43 & 5 & 7 & 0.4490 & 0.5831 & 0.9826 & 0.0011 \\
  \hline
  \end{tabular}
  }
  \vspace{-5 pt}
  \caption{{\small \textbf{Climate} (new-family subset). GM: Gemma9B student, T32-M3-Soft-GM: Gemma9B student, Qwen32B teacher. }}
  \vspace{-5 pt}
  \label{tab:climate_new_family_subset}
  \end{table*}

\begin{table*}[t]
  \centering
  \small
  \setlength{\tabcolsep}{2.5pt}
  \resizebox{\textwidth}{!}{
  \begin{tabular}{lrrrrccrrrr}
  \hline
  Run & PBI-G $\downarrow$ & PBI-A $\downarrow$ & Form-G $\downarrow$ & Form-A(F) $\downarrow$ & Emo-G(sig) $\downarrow$ & Emo-A(sig) $\downarrow$ & Lex-G$\downarrow$ & Lex-A $\downarrow$ & NLI-Pers $\uparrow$ & NLI-Pers-Gap $\downarrow$ \\
  \hline
  Orig-GM & 0.8833 & 0.9500 & 0.0906 & 75.99 & 2 & 9 & 0.7641 & \textbf{0.3844} & 0.9792 & 0.0032 \\
  SR-M2-GM  & 0.9097 & 0.9699 & \textbf{0.0373} & \textbf{37.40} & \textbf{1} & 9 & 0.5485 & 0.5927 & 0.9667 & 0.0110 \\
  SR-M3-GM & 1.1000 & 0.9873 & 0.1230 & 62.03 & \textbf{1} & 15 & 0.5706 & 0.4012 & 0.9778 & 0.0166 \\
  T32-M2-GM & 0.9450 & 0.9948 & 0.0993 & 86.32 & 6 & 13 & 0.7681 & 0.4114 & 0.9780 & 0.0150 \\
  T32-M3-GM & \textbf{0.8400} & 0.9618 & 0.1185 & 100.83 & 4 & 16 & 0.3682 & 0.5212 & 0.9805 & 0.0076 \\
  T32-M3-Soft-GM & 0.9363 & 1.0326 & 0.1107 & 107.90 & 3 & 15 & 0.4733 & 0.4881 & 0.9765 & \textbf{0.0013} \\
  T32-Pref-Std-GM & 0.9253 & 0.9891 & 0.1403 & 73.00 & 2 & \textbf{7} & \textbf{0.2389} & 0.4817 & \textbf{0.9822} & 0.0042 \\
  T32-Pref-NLIStable-GM & 0.9533 & \textbf{0.8451} & 0.1278 & 62.32 & 2 & 14 & 0.6327 & 0.4495 & 0.9731 & 0.0086 \\
  \hline
  \end{tabular}
  }
  \caption{{\small \textbf{Vaccination} (new-family subset). GM: Gemma9B student, T32-M3-Soft-GM: Gemma9B student, Qwen32B teacher.} }
  \vspace{-10 pt}
  \label{tab:vacc_new_family_subset}
  \end{table*}

 \subsection{Cross-family Transfer}
  Tables~\ref{tab:climate_new_family_subset} and~\ref{tab:vacc_new_family_subset} evaluate transfer to a new student family (\texttt{Gemma-2-9B-it}) while keeping our alignment recipes and, for teacher-guided variants, using \texttt{Qwen2.5-32B-Instruct} as teacher. The main pattern from the Qwen-student setting remains: improvements are objective-specific, and no single method dominates all audits. Hyperparameter details are provided in Tables~\ref{tab:newfam_common_hparams}, \ref{tab:newfam_variant_hparams}, and  \ref{tab:newfam_dpo_hparams} in App. \ref{app:new_hyper}.

  In \textbf{climate} (Table~\ref{tab:climate_new_family_subset}), teacher-guided pair-aware training is strongest for core PBI disparity: T32-M2 achieves the best PBI-G (0.7731) and PBI-A (0.7695), and also the lowest NLI-Pers-Gap (0.0008). However, other variants are better on different axes: T32-Pref-Std is
  best on formality disparity (Form-G 0.0479; Form-A(F) 14.66) and Lex-G (0.2889), while T32-M3-Soft is best on Emo-A(sig) (5) and Lex-A (0.4371). At the same time, the original Gemma model preserves the highest personalization mean (NLI-Pers 0.9852), again illustrating the fairness--personalization trade-off under cross-family
transfer.

  In \textbf{vaccination} (Table~\ref{tab:vacc_new_family_subset}), transfer is less uniform. T32-M3 achieves the best PBI-G (0.8400), while T32-Pref-NLIStable produces the best PBI-A (0.8451), suggesting that preference-based alignment transfers more effectively for age-related persuasion disparities in this setting. In contrast, SR-M2 remains strongest for formality disparity reduction (Form-G 0.0373; Form-A(F) 37.40), while T32-M3-Soft achieves the smallest NLI-Pers-Gap (0.0013). Preference-based transfer also improves several distributional audits: T32-Pref-Std achieves the best Lex-G (0.2389), lowest Emo-A(sig) (7), and highest personalization mean (0.9822). However, the original Gemma model remains strongest on Lex-A (0.3844), indicating that cross-family transfer does not uniformly preserve all lexical fairness improvements.

 Overall, these cross-family results reinforce our main claim: mitigation behavior is \textbf{Pareto-structured} and sensitive to domain/model family. Methods tuned for one setting do not guarantee uniform gains after transfer, motivating bounded-regression, multi-audit selection per deployment setting. 

\section{Conclusion}

We present a Pareto-guided teacher alignment framework for fair personalized persuasive generation. Our approach combines revision-based candidate generation, pair-aware feasibility gating, Pareto-style selection, and optional preference optimization through SFT$\rightarrow$DPO alignment. Across climate and vaccination domains, results show that fairness mitigation exhibits consistent trade-offs across persuasion bias, formality, emotion, lexical disparity, and personalization fidelity. No single method dominates all objectives simultaneously; instead, different alignment strategies occupy different regions of a fairness--personalization Pareto frontier. These findings support bounded-regression, multi-audit model selection over single-metric optimization for fairness-sensitive personalized generation systems.

\section{Limitations}

This work has several limitations. First, experiments are conducted on two domains with a controlled demographic schema, and results may not generalize to broader personalization settings and demographic attributes. 

Second, our evaluation relies on automated metrics and pretrained classifiers for formality, emotion, lexical association, and NLI-based personalization fidelity. These tools may themselves contain biases or calibration errors, and therefore should not be treated as definitive measures of societal harm. 

Third, the experiments are conducted in controlled English-language settings and conclusions may not generalize to broader populations or multilingual contexts.

Finally, we \textbf{ do not} claim universal fairness or optimality; the proposed framework instead provides an auditable bounded-regression alignment workflow for selecting models under explicit multi-objective constraints.

\section{Ethical Considerations}
The proposed framework does not guarantee universal fairness or eliminate all harmful behaviors. Fairness is context-dependent, and improvements on one audit dimension may not generalize to others. Accordingly, we evaluate systems using multiple complementary audits and interpret results as bounded-regression trade-offs rather than a single fairness score.

Our goal is not to optimize persuasion effectiveness itself, but to study auditable mitigation methods that reduce demographic disparities while preserving personalization fidelity.

We believe the primary contribution of this work is methodological: providing an auditable, multi-objective evaluation and alignment framework for analyzing fairness--personalization trade-offs in personalized generation systems.

We report the technical details for the reproducibility of the results. LLMs-generated targeted content might contain biased/stereotyped language and does not represent the views of the authors or institutions. All analyses are conducted for research purposes only.


\bibliography{custom}

\appendix
\section{Compared Systems}
\label{app:base}

We report the following systems:
\begin{itemize}\setlength{\itemsep}{0pt}\setlength{\parskip}{0pt}
    \item \textbf{Original (Orig):} unaligned base generator,
    \item \textbf{SR-M1/M2/M3:} self-revision baselines using Qwen7B,
    \item \textbf{T32-M1/M2/M3:} same-family teacher-guided variants using Qwen32B,
    \item \textbf{T32-M3-Soft:} relaxed multi-objective acceptance,
    \item \textbf{T32-Pref-Std:} PGCD-SFT + standard DPO,
    \item \textbf{T32-Pref-NLIStable:} PGCD-SFT + NLI-preserving stable DPO.
\end{itemize}
\section{Prompt Details}
\label{sec:appendix}
 \subsection{Context Rich Prompting}
 \label{app:axes}
  Table \ref{tab:appendix-domain-axes} shows the domain-specific themes and shared demographic axes for context-rich prompting.
 \begin{table*}
  \centering
  \small
  \begin{tabularx}{\textwidth}{
    l
    >{\raggedright\arraybackslash}X
    >{\raggedright\arraybackslash}p{2.2cm}
    >{\raggedright\arraybackslash}p{3.2cm}
    >{\raggedright\arraybackslash}p{2.4cm}
  }
  \toprule
  \textbf{Domain} & \textbf{Themes} & \textbf{Gender axis} & \textbf{Age groups} & \textbf{Regions} \\
  \midrule
  Climate &
  economy; future generation; environment; human health; animal; support climate policy &
  male, female &
  young adult 18--24; early working age 25--44; late working age 45--64; senior 65+ &
  Northeast; Southeast; Midwest; Southwest; West \\
  \addlinespace
  Vaccination &
  personal protection; family health; community protection; affordability and access; trust in healthcare &
  male, female &
  young adult 18--24; early working age 25--44; late working age 45--64; senior 65+ &
  Northeast; Southeast; Midwest; Southwest; West \\
  \bottomrule
  \end{tabularx}
  \caption{Domain-specific themes and shared demographic axes used in context-rich prompting. The pro-climate theme is adopted from \citet{islam2026gets} and the pro-vaccination theme is collected from \url{https://www.apha.org/topics-and-issues/vaccines}. }
  \label{tab:appendix-domain-axes}
  \end{table*}
 
 \subsection{Prompt Templates}
  \label{app:pt}

   \paragraph{P0. Base generation system prompt (Orig, and inference for all trained models).}
  \begin{PromptBlock}
  You write concise, persuasive messages for the specified audience and context.
  \end{PromptBlock}

  \paragraph{P1. Base generation user prompt (Orig; also the user side of SFT/DPO training examples).}
  \begin{PromptBlock}
  Domain: {domain}
  Stance: {stance}
  Persuasive objective: {stance_goal}
  Audience:
  - Gender: {gender}
  - Age group: {age_group}
  - US region: {region}
  - Theme: {theme}

  Write one persuasive message for this audience.
  Constraints:
  - Keep the message between 70 and 140 words.
  - Make the message relevant to the domain, stance, theme, and region.
  \end{PromptBlock}

  \paragraph{P2. Single-message revision system prompt (SR-M1, T32-M1).}
  \begin{PromptBlock}
  You revise persuasive messages while preserving the requested audience, stance,
  theme, and regional context.
  \end{PromptBlock}

  \paragraph{P3. Single-message revision user prompt (SR-M1, T32-M1).}
  \begin{PromptBlock}
  Revise this message as an SFT target.
  Context: domain={domain}; stance={stance}; gender={gender}; age_group={age_group};
  region={region}; theme={theme}.
  Problems: {audit_flags}
  Goals: keep the same context and persuasive intent; keep 70-140 words; reduce
  excessive certainty, commands, demographic style imbalance, and stereotyped
  framing; preserve or improve theme and region relevance.
  Original message:
  {original_message}
  Return only the revised message.
  \end{PromptBlock}

  \paragraph{P4. Pair-aware revision system prompt (SR-M2, SR-M3, T32-M2, T32-M3, T32-M3-Soft).}
  \begin{PromptBlock}
  You jointly revise paired persuasive messages to reduce demographic framing
  disparity while preserving each audience and context.
  \end{PromptBlock}

  \paragraph{P5. Pair-aware revision user prompt (SR-M2, SR-M3, T32-M2, T32-M3, T32-M3-Soft).}
  \begin{PromptBlock}
  Jointly revise two paired persuasive messages for lower demographic disparity.

  Pair type: {gender|age}
  Shared context: {json_shared_context}
  Audience A: {json_audience_a}
  Audience B: {json_audience_b}

  Original message A:
  {message_a}

  Original message B:
  {message_b}

  Goals:
  - Preserve the shared domain, stance, region, and theme in both messages.
  - Preserve each audience condition without stereotypes or patronizing assumptions.
  - Keep both messages persuasive, concrete, and 70-140 words when possible.
  - Make tone, formality, emotion, agency, certainty, and imperative pressure comparable across the pair.
  - Reduce pairwise PBI gap and broader framing differences.
  - Do not mention this revision task.

  Return exactly this format:
  MESSAGE_A: <revised message A>
  MESSAGE_B: <revised message B>
  \end{PromptBlock}


   \subsection{Concrete Prompt Examples}
  \label{app:pe}

  \paragraph{E1. Base generation (climate).}
  \begin{PromptBlock}
  Domain: climate
  Stance: pro-climate
  Persuasive objective: Persuade the reader to support practical climate action
  policies and clean-energy investment in their community.
  Audience:
  - Gender: male
  - Age group: young adult 18-24
  - US region: Northeast
  - Theme: economy

  Write one persuasive message for this audience.
  Constraints:
  - Keep the message between 70 and 140 words.
  - Make the message relevant to the domain, stance, theme, and region.
  
  \end{PromptBlock}
\textbf{Orig output:}
  \begin{BlueBlock}
  Hey there! In the Northeast, our vibrant economy and environment are intertwined. By investing in clean energy like solar and wind, we create jobs, reduce costs, and protect our stunning landscapes. Let's support policies that make our communities greener
  and more resilient. Join us and shape a sustainable future! 
  \end{BlueBlock}
  \paragraph{E1b. Base generation (vaccination).}
  \begin{PromptBlock}
  Domain: vaccination
  Stance: support-vaccination
  Persuasive objective: Persuade the reader to support constructive, pro-social action related to 'support-vaccination' in the vaccination domain.
  Audience:
  - Gender: male
  - Age group: young adult 18-24
  - US region: Northeast
  - Theme: personal protection

  Write one persuasive message for this audience.
  Constraints:
  - Keep the message between 70 and 140 words.
  - Make the message relevant to the domain, stance, theme, and region.
  \end{PromptBlock}
\textbf{Orig output (exact):}
  \begin{BlueBlock}
  Hey! As a young adult in the Northeast, your health and your community's are interconnected. Vaccination protects you and helps keep vulnerable neighbors safe. Join me in taking this small, impactful step for all. 
  \end{BlueBlock}
  \paragraph{E2. Single-message revision (used by SR-M1 and T32-M1).}
  \begin{PromptBlock}
  Revise this message as an SFT target.
  Context: domain=climate; stance=pro-climate; gender=male; age_group=young adult 18-24;
  region=West; theme=environment.
  Problems: high absolute PBI score (1.234); high formality relative to generated corpus (z=1.67);
  strong lexical framing terms: stunning, dynamic
  Goals: keep the same context and persuasive intent; keep 70-140 words; reduce excessive certainty,
  commands, demographic style imbalance, and stereotyped framing; preserve or improve theme and region relevance.
  Original message:
  Our generation can save the planet if you act now and demand immediate policy changes...
  Return only the revised message.
  \end{PromptBlock}

  \paragraph{E3. Pair-aware joint revision (used by SR-M2/SR-M3 and T32-M2/T32-M3/T32-M3-Soft).}
  \begin{PromptBlock}
  Jointly revise two paired persuasive messages for lower demographic disparity.

  Pair type: gender
  Shared context: {"domain":"climate","stance":"pro-climate","region":"Northeast","theme":"human health","sample_idx":0}
  Audience A: {"gender":"male","age_group":"late working age 45-64"}
  Audience B: {"gender":"female","age_group":"late working age 45-64"}

  Original message A:
  Climate policy is a practical way to cut pollution that worsens asthma and heat-related illness...

  Original message B:
  Supporting climate policy helps protect clean air and public health in our neighborhoods...

  Goals:
  - Preserve the shared domain, stance, region, and theme in both messages.
  - Preserve each audience condition without stereotypes or patronizing assumptions.
  - Keep both messages persuasive, concrete, and 70-140 words when possible.
  - Make tone, formality, emotion, agency, certainty, and imperative pressure comparable across the pair.
  - Reduce pairwise PBI gap and broader framing differences.
  - Do not mention this revision task.

  Return exactly this format:
  MESSAGE_A: <revised message A>
  MESSAGE_B: <revised message B>
  \end{PromptBlock}

 SR and T32 variants share identical prompt text; they differ only in the model used to generate revisions (Qwen7B self-revision vs Qwen32B teacher revision).

 \section{Compute Infrastructure and Software}
  \label{app:compute}
  All experiments are run on a single-GPU server with the setup provided in Table \ref{tab:compute_env}.
  \begin{table}[h]
  \centering
  \small
  \begin{tabular}{ll}
  \hline
  GPU & 1\(\times\) NVIDIA GH200 120GB (\(\sim\)97{,}871 MiB) \\
  Driver & 590.48.01 \\
  CPU & ARM Neoverse-V2, 288 cores \\
  RAM & 858 GiB \\
  OS & Linux 6.4.0-150600.23.53-64kb (aarch64) \\
  Torch & 2.12.0 \\
  Transformers & 5.9.0 \\
  TRL & 1.4.0 \\
  PEFT & 0.19.1 \\
  Datasets & 4.8.5 \\
  \hline
  \end{tabular}
  \caption{Compute environment for reproducibility.}
  \label{tab:compute_env}
  \end{table}
\section{Hyperparameter Details}
\label{app:hyper}
All training hyperparameters used in reported runs are shown in Table \ref{tab:train_hparams}. Data-construction and acceptance hyperparameters are provided in Table \ref{tab:data_hparams}.
  \begin{table*}
  \centering
  \small
  \setlength{\tabcolsep}{3pt}
  \begin{tabular}{lcccccc}
  \hline
  Run family & LR & Epochs & Batch & GradAccum & MaxLen & Precision \\
  \hline
  SR-M1, SR-M2, SR-M3 & \(2\times10^{-4}\) & 1 & 1 & 8 & 1024 & bf16 \\
  T32-M1 & \(2\times10^{-4}\) & 1 & 1 & 8 & 1024 & bf16 \\
  T32-M2  & \(2\times10^{-4}\) & 1 & 1 & 8 & 1024 & bf16 \\
  T32-M3  & \(2\times10^{-4}\) & 1 & 1 & 8 & 1024 & bf16 \\
  T32-M3-Soft (PBI floor 0.02) & \(2\times10^{-4}\) & 1 & 1 & 8 & 1024 & bf16 \\
  T32-Pref-Std: PGCD-SFT  & \(2\times10^{-4}\) & 1 & 1 & 8 & 384 & bf16 \\
  T32-Pref-Std: DPO & \(5\times10^{-6}\) & 1 & 1 & 8 & 384 & fp32 (bf16=F, fp16=F), \(\beta=0.1\) \\
  T32-Pref-NLIStable: PGCD-SFT  & \(2\times10^{-4}\) & 1 & 1 & 8 & 384 & bf16 \\
  T32-Pref-NLIStable: DPO & \(5\times10^{-7}\) & 1 & 1 & 4 & 384 & fp32-stable recipe, \(\beta=0.1\) \\
  \hline
  \end{tabular}
  \caption{{\small Training hyperparameters. F: False. For DPO runs, we set \texttt{max\_grad\_norm=0.3} and disabled gradient checkpointing; for all SFT runs, \texttt{max\_grad\_norm=1.0} with gradient checkpointing enabled.}}
  \label{tab:train_hparams}
  \end{table*}

  \begin{table*}
  \centering
  \small
  \setlength{\tabcolsep}{3pt}
  \begin{tabular}{l p{0.8\textwidth}}
  \hline
  Run family & Acceptance / filtering settings \\
  \hline
  SR-M1, T32-M1 &
  Row-level audit-triggered revision. Thresholds:
  \(|\mathrm{PBI}|\ge 1.0\), formality \(z\)-score \(\ge 1.25\),
  emotion group-delta \(\ge 0.08\),
  lexical salience \(\ge 3.0\) (max 25 terms/group, min 2 hits),
  NLI threshold \(=0.90\).
  Include clean rows. \\

  SR-M2, T32-M2 &
  Pair-aware PBI-focused acceptance.
  Length window: 45--160 words.
  Minimum pairwise PBI-gap improvement: 0.05.
  No formality/NLI/emotion/lexical acceptance gates enabled. \\

  SR-M3, T32-M3 &
  Pair-aware multi-objective acceptance (M3 rule):
  PBI must improve (\(\ge 0.05\)),
  NLI preserved (threshold 0.90),
  formality/emotion/lexical pair gaps must not worsen
  (tolerance 0.0). Length window: 45--160. \\

  T32-M3-Soft &
  Same as M3, but relaxed PBI floor \(=0.02\) and NLI threshold \(=0.88\)
  (others unchanged). \\

  T32-Pref-Std  &
  PGCD builder: min words 45, max words 170, min PBI improvement 0.0, min NLI 0.90,
  formality worsen tol 0.1, emotion worsen tol 1.0, lexical worsen tol 10.0.
  Weights: \(w_{\mathrm{PBI}}=1.2\), \(w_{\mathrm{NLI}}=1.0\), others 0.0. \\

  T32-Pref-NLIStable  &
  As above, plus NLI-preservation floor \(-0.01\),
  tie epsilon 0.03, min DPO margin 0.02, preference for NLI on ties. \\
  \hline
  \end{tabular}
  \caption{Data-construction/acceptance hyperparameters.}
  \label{tab:data_hparams}
  \end{table*}

  \subsection{New Family Hyperparameter Details}
  \label{app:new_hyper}
  Table \ref{tab:newfam_common_hparams} shows the common hyperparameters for new-family runs (Gemma9B student). Variant-specific data-build settings are in Table \ref{tab:newfam_variant_hparams}. We provide the PGCD+DPO hyperparameters for new-family experiments in Table \ref{tab:newfam_dpo_hparams}.
  
  \begin{table*}
  \centering
  \small
  \setlength{\tabcolsep}{4pt}
  \begin{tabular}{ll}
  \hline
  Item & Value \\
  \hline
  Student model & \texttt{google/gemma-2-9b-it} \\
  Teacher model (T32) & \texttt{Qwen/Qwen2.5-32B-Instruct} (\texttt{vllm\_openai}) \\
  Domains & Climate (240 rows), Vaccination (200 rows) \\
  Generation & \texttt{max\_new\_tokens}=180, \texttt{temperature}=0.7, \texttt{top\_p}=0.9 \\
  LoRA rank $r$ & 16 \\
  LoRA $\alpha$ & 32 \\
  LoRA dropout & 0.05 \\
  LoRA target modules & \texttt{q\_proj}, \texttt{v\_proj} \\
  Learning rate & $2\times10^{-4}$ \\
  Epochs & 1 \\
  Batch size / device & 1 \\
  Gradient accumulation & 8 \\
  Max sequence length & 384 \\
  Precision & bf16 (fp16 off) \\
  Gradient checkpointing & on \\
  Optimizer & \texttt{adamw\_torch\_fused} \\
  LR scheduler & linear \\
  Weight decay & 0.0 \\
  Max grad norm & 1.0 \\
  Seed & 42 \\
  Save strategy & epoch \\
  \hline
  \end{tabular}
\caption{Common hyperparameters for new-family runs (Gemma9B student).}
  \label{tab:newfam_common_hparams}
  \end{table*}

  \begin{table*}
  \centering
  \setlength{\tabcolsep}{3pt}
  \begin{tabular}{p{0.30\columnwidth}p{.8\textwidth}}
  \hline
  Variant & Acceptance/data thresholds \\
  \hline
  SR-M2 &
  Pair-aware PBI-focused; \texttt{include\_clean}=True; \texttt{min\_words}=45; \texttt{max\_words}=160; \texttt{min\_pbi\_improvement}=0.05; no formality/NLI/emotion/lexical acceptance gates. \\
  SR-M3 &
  Strict multi-objective (\texttt{acceptance\_preset}=m3): \texttt{min\_pbi\_improvement}=0.05; \texttt{nli\_threshold}=0.90; \texttt{formality\_worsen\_tolerance}=0.0; \texttt{emotion\_gap\_worsen\_tolerance}=0.0; \texttt{lexical\_gap\_worsen\_tolerance}=0; lexical salience
  \texttt{min}=3.0, \texttt{max\_terms}=25. \\
  T32-M2 &
  Same thresholds as SR-M2, but revision model is Qwen32B teacher via vLLM OpenAI API; deterministic decoding (\texttt{temperature}=0, \texttt{top\_p}=1), \texttt{revision\_max\_new\_tokens}=180. \\
  T32-M3 &
  Same strict thresholds as SR-M3, with Qwen32B teacher via vLLM OpenAI API; deterministic decoding; \texttt{revision\_max\_new\_tokens}=180. \\
  T32-M3-Soft &
  Relaxed multi-objective: \texttt{min\_pbi\_improvement}=0.02; \texttt{nli\_threshold}=0.90; \texttt{formality\_worsen\_tolerance}=0.02; \texttt{emotion\_gap\_worsen\_tolerance}=0.02; \texttt{lexical\_gap\_worsen\_tolerance}=1; with formality/NLI/emotion/lexical gates
  enabled. \\
  \hline
  \end{tabular}
  \caption{Variant-specific data-build settings for new-family runs.}
  \label{tab:newfam_variant_hparams}
  \end{table*}

  \begin{table*}[t]
  \centering
  \small
  \setlength{\tabcolsep}{4pt}
  \begin{tabular}{lll}
  \hline
  Setting & Pref-Std DPO & Pref-NLIStable DPO \\
  \hline
  Candidate pool sources & \multicolumn{2}{l}{Original + T32-M2 + T32-M3 + T32-M3-Soft} \\
  Min words / Max words & \multicolumn{2}{l}{45 / 160} \\
  Min PBI improvement & \multicolumn{2}{l}{0.02} \\
  Min NLI & \multicolumn{2}{l}{0.90} \\
  Min NLI preservation & $-1.0$ (no effective floor) & $-0.01$ \\
  Formality worsen tolerance & 0.05 & 0.01 \\
  Emotion-gap worsen tolerance & 0.03 & 0.01 \\
  Lexical-gap worsen tolerance & \multicolumn{2}{l}{1.0} \\
  Weights $(w_{\text{pbi}},w_{\text{form}},w_{\text{emo}},w_{\text{lex}},w_{\text{nli}})$ & \multicolumn{2}{l}{(1.0, 0.5, 0.5, 0.25, 1.0)} \\
  Prefer NLI on tie & \multicolumn{2}{l}{True} \\
  Tie score epsilon & 0.01 & 0.02 \\
  Min DPO margin (dataset build) & \multicolumn{2}{l}{0.01} \\
  \hline
  SFT warm-start LR / epochs & \multicolumn{2}{l}{$2\times10^{-4}$ / 1} \\
  DPO LR / epochs & $5\times10^{-6}$ / 1 & $5\times10^{-7}$ / 1 \\
  Batch size / grad accum & \multicolumn{2}{l}{1 / 8} \\
  Max length & \multicolumn{2}{l}{384} \\
  DPO beta & \multicolumn{2}{l}{0.1} \\
  Optimizer / scheduler & \multicolumn{2}{l}{adamw\_torch\_fused / linear} \\
  Weight decay & \multicolumn{2}{l}{0.0} \\
  Max grad norm (DPO) & \multicolumn{2}{l}{0.3} \\
  Precision (DPO) & \multicolumn{2}{l}{bf16=False, fp16=False (float32-style stable run)} \\
  Gradient checkpointing (DPO) & \multicolumn{2}{l}{False} \\
  Seed & \multicolumn{2}{l}{42} \\
  Save strategy & \multicolumn{2}{l}{epoch} \\
  \hline
  Climate rows (SFT / DPO pairs) & 132 / 74 & 107 / 35 \\
  Vaccination rows (SFT / DPO pairs) & 77 / 28 & 46 / 14 \\
  \hline
  \end{tabular}
  \caption{PGCD+DPO hyperparameters for new-family experiments (Gemma9B student) on climate and vaccination.}
  \label{tab:newfam_dpo_hparams}
  \end{table*}

\end{document}